%% file: root.tex
\documentclass[letterpaper, 10 pt, conference]{ieeeconf}  % Comment this line out if you need a4paper

\IEEEoverridecommandlockouts                              % This command is only needed if 
\usepackage{comment}
\usepackage{amsmath,amssymb}
\usepackage{color}
\usepackage{url}
\usepackage{hyperref}

\usepackage{acronym}

\acrodef{6dof}[6DoF]{six degrees of freedom}
\acrodef{ad}[AD]{autonomous driving}
\acrodef{av}[AV]{autonomous vehicle}
\acrodef{cnn}[CNN]{convolutional neural network}
\acrodef{dnn}[DNN]{deep neural network}
\acrodef{imu}[IMU]{inertial measurement unit}
\acrodef{iou}[IoU]{intersection-over-union}
\acrodef{gnss}[GNSS]{Global Navigation Satellite System}
\acrodef{gps}[GPS]{Global Positioning System}
\acrodef{lidar}[LiDAR]{light detection and ranging}
\acrodef{miou}[mIoU]{mean \ac{iou}}
\acrodef{mvs}[MVS]{multiple-view stereo}
\acrodef{oa}[OA]{overall accuracy}
\acrodef{ac}[mAcc]{mean accuracy}
\acrodef{rtk-gnss}[RTK-GNSS]{real-time kinematic \ac{gnss}}
\acrodef{slam}[SLAM]{simultaneous localization and mapping}
\acrodef{sfm}[SfM]{structure-from-motion}
\acrodef{vo}[VO]{visual odometry}
\acrodef{fcnn}[FCNN]{fully convolutional neural network}
\acrodef{crf}[CRF]{Conditional Random Field}
\acrodef{tcl}[TCL]{temporally consistent labeling}

\acused{lidar}
\acused{cnn}
\acused{slam}

\def\rot{\rotatebox[origin=lb]}
\usepackage{lscape}
\usepackage{multirow}
\usepackage{graphicx}
\usepackage{float}
\usepackage{placeins}
\usepackage{siunitx}

\usepackage{etoolbox}
\makeatletter
\patchcmd{\@makecaption}
  {\scshape}
  {}
  {}
  {}
\makeatother

\usepackage{caption}
\usepackage{comment}
\usepackage{lipsum}
\title{\LARGE \bf
Vision-based Large-scale 3D Semantic Mapping for Autonomous Driving Applications}

\author{
\authorblockN{Qing Cheng$^{1}$ \authorblockA{qing@artisense.ai}} \and
\authorblockN{Niclas Zeller$^{1,3}$ \authorblockA{niclas.zeller@h-ka.de}} \and
\authorblockN{Daniel Cremers$^{1,2}$ \authorblockA{cremers@tum.de}}
\thanks{$^{1}$ Artisense GmbH}%
\thanks{$^{2}$ Technical University of Munich}%
\thanks{$^{3}$ Karlsruhe University of Applied Sciences}%
}

\begin{document}

\maketitle

%%%%%%%%%%%%%%%%%%%%%%%%%%%%%%%%%%%%%%%%%%%%%%%%%%%%%%%%%%%%%%%%%%%%%%%%%%%%%%%%
\begin{abstract}
In this paper, we present a complete pipeline for 3D semantic mapping solely based on a stereo camera system.
The pipeline comprises a direct sparse visual odometry front-end as well as a back-end for global optimization including GNSS integration, and semantic 3D point cloud labeling.
We propose a simple but effective temporal voting scheme which improves the quality and consistency of the 3D point labels.
Qualitative and quantitative evaluations of our pipeline are performed on the KITTI-360 dataset.
The results show the effectiveness of our proposed voting scheme and the capability of our pipeline for efficient large-scale 3D semantic mapping.
The large-scale mapping capabilities of our pipeline is furthermore demonstrated by presenting a very large-scale semantic map covering 8000 km of roads generated from data collected by a fleet of vehicles.
\end{abstract}%


%%%%%%%%%%%%%%%%%%%%%%%%%%%%%%%%%%%%%%%%%%%%%%%%%%%%%%%%%%%%%%%%%%%%%%%%%%%%%%%%

\input{sections/introduction}
\input{sections/related_work}
\input{sections/method}
\input{sections/experiments}
\input{sections/conclusion}

\input{sections/acknowledgement}
% \FloatBarrier

\bibliographystyle{IEEEtran}
\bibliography{references}

\end{document}

%% file: sections/introduction.tex
\section{Introduction}\label{sec:introduction}

Autonomous Driving (AD)\acused{ad} is among the major technical challenges pursued by automotive companies and academics around the globe.
One of the most considerable obstacles on the way to full autonomy is the challenge of 3D perception: understanding the 3D world around the vehicle with its various sensors.
An \ac{av} can perform online path planning and  make reliable, safety-critical decisions based on such an established environmental model.

While in the past reliable perception was enabled by equipping a vehicle with all different kinds of sensor modalities, today's trends go more towards the use of fewer and cheaper sensors to perceive the surrounding of a vehicle.

State of today, in addition to online perception, environmental models are complemented by topological information of static road furniture.
High-definition maps (HD maps) can provide such redundant and abundant information to back up the online sensor data.
Nevertheless, it is crucial to keep such maps up to date due to rapid changes in road infrastructures, especially in urban environments.
Hence, instead of expensive mapping sensors and manual labeling processes, lightweight and scalable online mapping pipelines become more preferred.

\begin{figure}[t]
    \centering
    \includegraphics[width=0.49\textwidth]{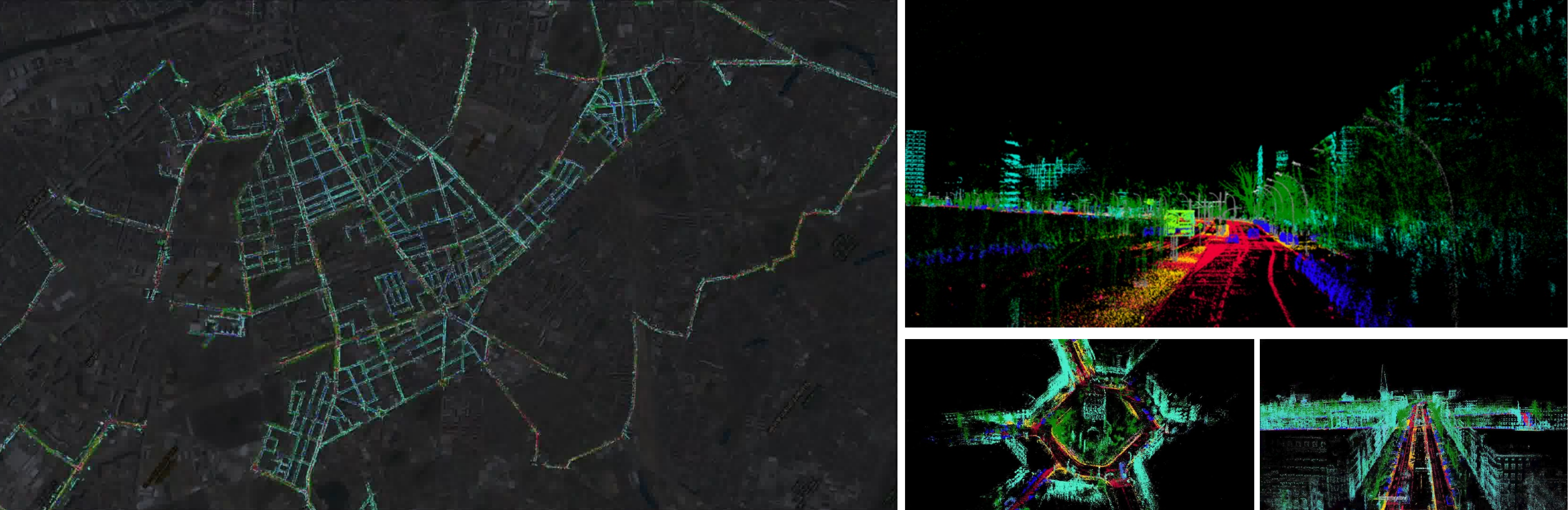}
    \vspace{-1em}
    \caption{\textbf{Large-scale semantic map of Berlin.} \textbf{\textit{Left:}} We generated a large-scale semantic map covering \SI{8000}{km} of roads in Berlin with a fleet of vehicles. \textbf{\textit{Right:}} Zoomed-in sections of the semantic map show the fine-grained 3D reconstruction details of the map.}
    \label{fig:berlin_map}
\end{figure}

\begin{figure}[t]
    \centering
    \includegraphics[width=0.49\textwidth]{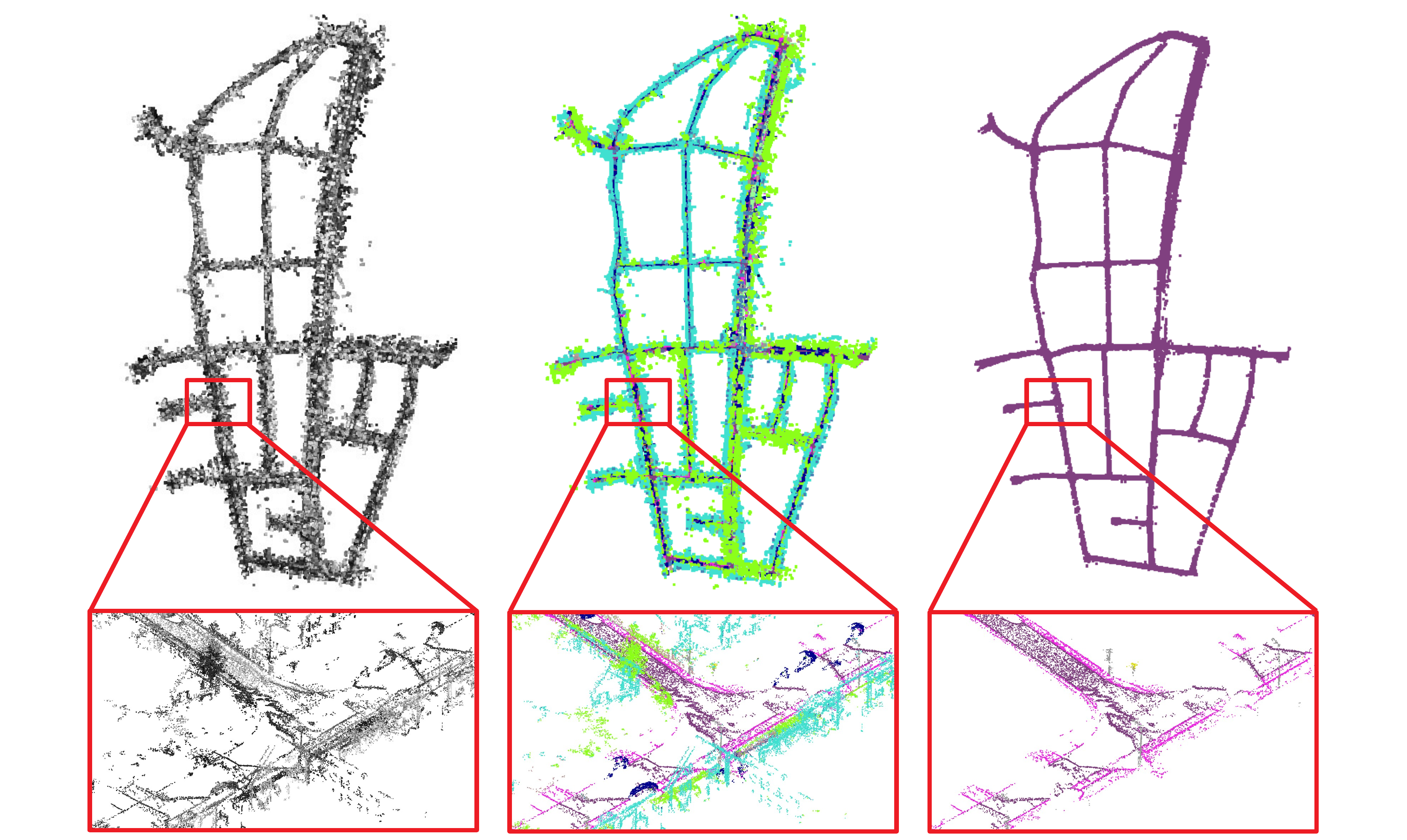}
    \vspace{-1em}
    \caption{\textbf{3D semantic map} generated by the proposed pipeline for sequence 0 of the KITTI-360 dataset~\cite{xie2016kitti360}. \textbf{\textit{Left:}} Sparse point cloud generated from direct \ac{vo}. \textbf{\textit{Center:}} Semantic 3D point cloud. \textbf{\textit{Right:}} Extracted street-level information, including roads, sidewalks, traffic signs/lights. Due to the high accuracy and quality, the generated semantic map can be used to perform further vectorization towards HD map generation.}
    \label{fig:semantic_map_sample}
\end{figure}

In this paper, we are proposing a complete vision-based pipeline towards the goal of creating scalable and up-to-date maps.
Our pipeline generates 3D semantic maps at scale from only a stereo-vision system, as shown in Figure~\ref{fig:semantic_map_sample}.
We believe that the proposed pipeline reveals the potential of purely vision-based mapping systems for \acl{ad} applications and can be extended towards extracting information like lane markings, etc., although it does not provide fully vectorized HD maps yet.
In Figure~\ref{fig:berlin_map}, we demonstrate that our method can be used to create city-scale maps based on a fleet of vehicles.

\textbf{Our Contributions} in detail are the following:
\begin{itemize}
    \item A fully automatic vision-based 3D mapping pipeline that can create large-scale 3D semantic maps efficiently.
    \item A simple but effective temporally consistent point labeling scheme that improves the accuracy of the 3D point labels by taking the structural information provided by the \acl{vo} front-end into account.
    \item A benchmark for vision-based 3D semantic mapping pipelines, which fuses 3D \ac{lidar} and 2D image ground truth labels. 
\end{itemize}

%% file: sections/related_work.tex
\section{Related Work}\label{sec:related_work}

In this paper, we are presenting a pure vision-based semantic mapping pipeline.
Therefore, there are basically two different research streams to which our method relates.
One is visual \ac{slam}, which allows building large-scale 3D maps only from image data.
The other is semantic segmentation, which aims to extract semantic information from the captured sensor data.
Hence, in the sequel, we review both the state of the art in visual \ac{slam} as well as in 2D (image-based) and 3D (\ac{lidar}-based) semantic segmentation.
Furthermore, we discuss related mapping pipelines based on different sensor modalities. 
For all these tasks, there exists several popular datasets relevant for training and benchmarking~\cite{geiger2012kitti,cordts2016cityscapes,xie2016kitti360,neuhold2017mapillary,behley2019semanticKitti,wang2019apolloscape,wenzel2020fourseasons}.

\subsubsection{Visual \acs{slam}}
or \ac{sfm} algorithms are at the core of any mapping pipeline.
These methods can be mainly divided into indirect (or keypoint-based) and direct approaches.

Indirect approaches extract a set of interest points from the image and then perform pose and 3D point estimation by formulating a geometric energy term using the extracted interest points \cite{klein2007ptam,murartal2015orbslam,murartal2017orb2,elvira2019orbatlas,usenko2020basalt,campos2021orb3}.
Here correspondences of interest points between consecutive images are either established by matching descriptors \cite{klein2007ptam,murartal2015orbslam} or based on optical flow estimation \cite{usenko2020basalt}.
While indirect methods generally provide high tracking robustness, the 3D reconstruction is limited to the sparse set of interest points, making it insufficient for 3D mapping purposes.
This can be overcome by combining an indirect pose estimation with \ac{mvs} \cite{murartal2015mapping,schoeps2015onthego,schoenberger2016mvs}.

Direct approaches skip the keypoint extraction step and perform optimization directly based on pixel intensities \cite{engel2014lsd,engel2018dso}.
As these methods generally make use of much more image points than indirect approaches, they can provide a richer and more detailed 3D reconstruction of the environment \cite{engel2014lsd,engel2018dso,wang2017stereoDSO}.
Direct methods do not establish any point-to-point correspondences between images as done in indirect approaches, so they tend to be less robust under very fast motion and rapidly changing lighting conditions.
In \cite{yang20d3vo}, this problem is tackled by integrating predictions from a \ac{dnn} into a direct \ac{slam} formulation.
While tasks like loop closure detection and relocalization, in general, are very challenging for direct approaches due to the non-repeatable selection of points, \cite{gao2018ldso,gladkova2021tightfusion} have shown that this bottleneck can be overcome by integrating detected keypoints into a direct \ac{slam} system.
In contrast to \ac{slam} approaches which are designed for online operation and sequential image data, \ac{sfm} tackles the problem in a more general way by performing 3D reconstruction from a set of unordered images \cite{schoenberger2016sfm,demmel2020distBA}.

\subsubsection{Semantic Segmentation}\label{sec:related_work_semseg}
is known as the task of assigning a semantic class label to each recorded data point (e.g. image pixel or 3D \ac{lidar} point).
In recent years, research in this field is heavily driven by large-scale datasets~\cite{cordts2016cityscapes,xie2016kitti360,neuhold2017mapillary,behley2019semanticKitti,wang2019apolloscape}.
Two main work streams are present for semantic segmentation in robotic and \acl{ad} use-cases: 2D semantic segmentation on RGB or intensity images and 3D semantic segmentation on structural 3D representation, e.g. \ac{lidar} point clouds.

Due to the high resolution of images and the rich textural information, image-based approaches provide high-quality segmentation results~\cite{long2015fcnsemseg,zhao2017pspn,chen2017deeplabv3}.
In general, these methods are highly efficient using 2D \acp{cnn}.
The pioneering work \cite{long2015fcnsemseg} tackles the semantic segmentation task with a \ac{fcnn} which can generate a full-resolution semantic mask for an input image of arbitrary size.
To utilize more high-level context, \cite{chen2015semantic,schwing2015fullycd,zheng2015crf,lin2016efficient} integrate the probabilistic graphical models, e.g. \acp{crf}, into their network architectures. Another direction is to exploit multi-scale inputs \cite{eigen2015predictingds,chen2016attentionts,farabet2013learninghfsl,pinheiro2014recurrentcn} or spatial pyramid pooling \cite{Chen2018DeepLabSI,chen2017deeplabv3,Zhao2017PyramidSP,Liu2015ParseNetLW} in order to generate feature maps encoding contextual information from different resolutions for the final pixel-level classification. The popular encoder-decoder architectures learn dense semantic features from the encoded low-resolution feature responses from single or multiple encoder layers and demonstrate promising performance on semantic segmentation tasks with this scale context \cite{Noh2015LearningDN,Vijay2017SegNetAD,Ronneberger2015UNetCN,Lin2017RefineNetMR,Pohlen2017FullResolutionRN,fu2017stackeddn}. Attention mechanisms are also introduced to semantic segmentation architectures which outshine max and average pooling with the selective aggregation of various contextual features \cite{chen2016attentionts,Fu2019DualAN,Huang2019CCNetCA,yuan2021ocnet,Tao2020HierarchicalMA,Yu2018LearningAD}. Auto-labelling techniques \cite{zhu2019improving,iscen2019label,Tao2020HierarchicalMA} and domain adaptation methods \cite{zou2018unsupervised,li2019bidirectional} are also proposed to alleviate the reliance on the expensive pixel-wisely labeled training data.
In our work, we make use of the work proposed by \cite{zhu2019improving}, which demonstrates state-of-the-art performance in automotive environments.

\ac{lidar}-based 3D segmentation approaches commonly rely on the geometric point distribution to predict semantic point labels.
One of the main challenges for these approaches is the unstructured nature of the 3D point cloud.
This challenge can be tackled by transforming the point cloud e.g. into a range image representation and applying classical 2D \acp{cnn}~\cite{wu2018squeezeseg,wang2018pointseg}.
To better utilize the 3D nature of point clouds, \cite{tchapmi2017segcloud} proposes to apply 3D \acp{cnn} to a voxel representation obtained from the point cloud.

\subsubsection{Semantic Mapping}
While the approaches presented in Section~\ref{sec:related_work_semseg} solely focus on the segmentation problem using individual images or individual \ac{lidar} scans, we understand the task of semantic mapping as the combination of segmentation, localization and mapping.

In \cite{sengupta2012semmapping}, dense 2D semantic mapping is performed by projecting class labels from street view images onto the ground plane.
Furthermore, in \cite{sengupta2013stereomapping} an extension to \cite{sengupta2012semmapping} is proposed, which performs dense reconstruction from stereo images and labels the resulting 3D model based on image segmentation results.

Instead of using only image data, the methods proposed in \cite{kostavelis2015mobilerobot} and \cite{maturana2017offroad} make use of a \ac{lidar} sensor for 3D map generation and perform semantic image segmentation to label the 3D points.
The work presented in \cite{paz2020semanticmapping} also makes use of \ac{lidar}-based 3D reconstructions and image-based segmentation to reconstruct both 3D semantic point clouds as well as semantic 2D occupancy grids.

%% file: sections/method.tex
\section{Large-scale 3D Semantic Mapping}\label{sec:method}

\begin{figure}[t]
    \centering
    \includegraphics[width=0.49\textwidth]{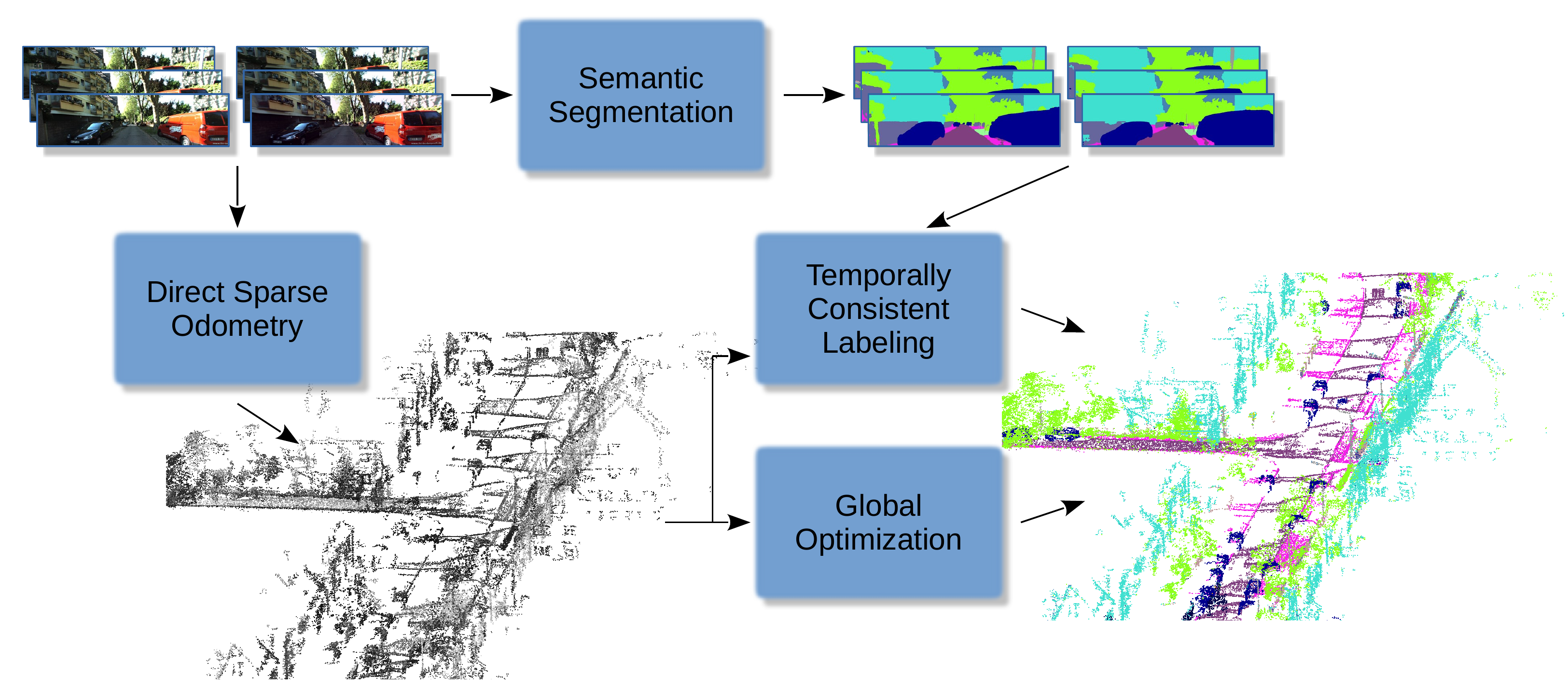}
    \vspace{-1.em}
    \caption{\textbf{Semantic Mapping Pipeline.}
    A direct \acs{vo} front-end estimates relative camera poses and a sparse 3D reconstruction of the environment based on a sequence of stereo images.
    A segmentation module predicts dense 2D semantic labels for the stereo camera images selected as keyframes by the \acs{vo}. 
    Based on the \acs{vo} outputs and the 2D semantic labels, temporally consistent 3D point labels are generated.
    Global optimization can be applied to compensate for drift based on loop closure detection and the integration of \acs{rtk-gnss} measurements if available.}
    \label{fig:pipeline}
\end{figure}

The proposed 3D semantic mapping system utilizes solely stereo images and optional sensor data like \acused{gnss}\acs{gnss} (\acl{gnss}) for global positioning and \acs{imu} measurements.
Figure~\ref{fig:pipeline} shows an overview of our semantic mapping pipeline.
The direct \ac{vo} module of our pipeline estimates relative camera poses and a sparse 3D reconstruction of the environment (Sec.~\ref{sec:3d_mapping}).
Global map optimization is performed based on loop closure detection.
If available, RTK-\ac{gnss} measurements are integrated to obtain a geo-referenced and globally accurate map.
A state-of-the-art semantic segmentation network (Sec.~\ref{sec:semantic_segmentation}) is used to generate accurate semantic labels for each pixel of the stereo images corresponding to the keyframes defined by the \ac{vo} front-end.
Furthermore, our temporally consistent labeling module can generate temporally consistent 3D point labels based on the \ac{vo} outputs and the 2D semantic labels (Sec.~\ref{sec:temporal_filtering}).
When lifting the 3D reconstruction to a global, geo-referenced, frame with \ac{gnss} data, a city-scale map can be created by stitching together the reconstructions from a fleet of vehicles, as shown in Figure~\ref{fig:berlin_map}.

\subsection{Visual Odometry and 3D Mapping}\label{sec:3d_mapping}
The core of the proposed semantic mapping pipeline is a state-of-the-art visual \ac{slam} algorithm.
To provide rich information about the 3D environment, a detailed reconstruction of the surroundings at a certain density level is required.
The proposed semantic mapping pipeline relies on a direct \ac{slam} front-end for 3D map generation, which provides a much denser and more detailed reconstruction of the environment compared to indirect (keypoint-based) \ac{slam} approaches.

Namely, we make use of a direct sparse \ac{slam} formulation working on stereo images, as proposed in \cite{wang2017stereoDSO}.
Here, in a local photometric bundle-adjustment, \ac{6dof} poses are estimated jointly with a sparse 3D reconstruction of the environment and optimized with respect to a photometric energy function:
\begin{align}
    E = \sum_{i \in \mathcal{F}}{\sum_{\boldsymbol{p} \in \mathcal{P}_i}{\left( \left(\sum_{j \in \text{obs}^t(\boldsymbol p)}{E_{ij}^{\boldsymbol p}}\right) + \lambda E_{is}^{\boldsymbol p}  \right)}}.
    \label{eq:photometric_energy}
\end{align}
In eq.~\eqref{eq:photometric_energy}, $\mathcal{F}$ is the set of all keyframes in the optimization window and $\mathcal{P}_i$ is the sparse set of points hosted in keyframe $i$.
A point $\boldsymbol p \in \mathcal{P}_i$ is defined by its fixed 2D pixel location in the image and an inverse depth $d$ which is estimated during optimization.
$\text{obs}^t(\boldsymbol p)$ defines the projections of the point $\boldsymbol p$ to the left images of all other keyframes in the optimization window.
Furthermore, $E_{is}^{\boldsymbol p}$ defines an additional photometric energy term based on the projection of a point $\boldsymbol p$ into the corresponding right image.
For further details, we refer to~\cite{engel2018dso,wang2017stereoDSO}.

Furthermore, our \ac{vo} system additionally is able to tightly integrate inertial measurements from an \ac{imu} if available, as described in \cite{stumberg2018vidso}.
By fusing the \ac{vo} outputs with \ac{gnss} measurements and performing loop closure detection, a globally consistent 3D map is obtained.
Overall, the mapping front-end used in our pipeline is similar to the one proposed in \cite{wenzel2020fourseasons}.

\subsection{Semantic Segmentation}\label{sec:semantic_segmentation}
Semantic segmentation plays an influential role in our pipeline as it sets the baseline for the quality of our semantic point clouds. Since there are already plentiful excellent works, we build our pipeline based on them.
Because of the modularity of our pipeline, the semantic segmentation component consumes only images and, thus, is independent of the rest of the pipeline. Therefore, any start-of-the-art method can be integrated into our pipeline.

In this paper, we build our pipeline based on \cite{zhu2019improving}.
The work proposed in~\cite{zhu2019improving} has the best performance on the KITTI semantic segmentation benchmark \cite{Alhaija2018IJCV} among all open-source submissions. It achieves state-of-the-art performance by augmenting the training set with synthesised samples generated by the joint image-label propagation model along with the boundary label relaxation strategy to make the model robust to noisy boundaries. 

In our pipeline, the semantic segmentation module predicts a full-resolution semantic map for each pair of images from the stereo cameras corresponding to the keyframes defined by the \ac{vo} module and then feeds the consecutive maps into the temporally consistent labeling module (Section~\ref{sec:temporal_filtering}).

\subsection{Temporally Consistent Labeling}\label{sec:temporal_filtering}
Even though \acp{dnn} for image-based semantic segmentation significantly improved over the last years, these networks still suffer from noise in the predicted labels as well as temporal inconsistency across the estimations of a temporal sequence of images.
On one side, there are usually significant appearance changes of objects caused by the perspective imaging process.
On the other side, the prediction capability of \acp{dnn} is still developing and imperfect estimations can often be temporal inconsistent.

However, with respect to the task of \acl{ad}, one is generally more interested in temporally consistent 3D semantic labels than 2D labels in image space.
Therefore, we are proposing a simple but effective scheme to obtain accurate and consistent 3D labels only from images.

As described in Section~\ref{sec:3d_mapping}, our direct sparse \ac{vo} system \cite{engel2018dso,wang2017stereoDSO} generates a sparse set of points\linebreak $\mathcal{P} := \{\mathcal{P}_0, \mathcal{P}_i, \cdots, \mathcal{P}_N\}$ to define the 3D reconstruction.
Here a point $\boldsymbol p \in \mathcal{P}_i$ is defined by the pixel coordinates $(u,v)$ in its host keyframe $i \in \mathcal{F}$ as well as a corresponding inverse depth $d$.
Therefore, a point can be projected to a global 3D frame using the corresponding keyframe pose $\boldsymbol \xi_i \in \mathfrak{se}(3)$.
In general, the tracked points are treated as statistically independent from each other.
Furthermore, the point selection scheme avoids having multiple image points hosted in different frames corresponding to the same object points.

In the segmentation process described in Section~\ref{sec:semantic_segmentation}, the individual images of the sequence are treated independent from each other.
While the direct \ac{vo} front-end implicitly establishes correspondences between consecutive images as defined in eq.~\eqref{eq:photometric_energy}.
Therefore, we can resort to these correspondences to improve the quality of the 3D point labels.

Similar to the photometric energy formulation in eq.~\eqref{eq:photometric_energy}, we project each point $\boldsymbol p \in \mathcal{P}_i$ from its host keyframe $i$ to the left images corresponding to all other keyframes $j$ in the neighborhood in which the point $\boldsymbol p$ is visible ($j \in \text{obs}^t(\boldsymbol p)$) to generate a set of co-visible points corresponding to this point.
We also project this set of left image points to their respective right images to obtain another set of corresponding points. Both sets of points from the left and right image stream constitute the final co-visible point set. 
Then we define a set of $C$ votes\linebreak $V := \{v_1, v_2, \cdots, v_{C}\}$ for each point $\boldsymbol p$, where $v_c$ is defined as an accumulated vote based on the weights $w_j$ from all co-visible points for one semantic class $c$ and $C$ is the total number of classes.
\begin{align}
\begin{split}
    v_c = \sum_{j \in \mathcal{Q}}{w_j}
\end{split}
\begin{split}
    w_j = \begin{cases}
        d_j & \text{for } c = L_j(\boldsymbol q_j) \text{ and } d_j < \text{dist}_{\min}^{-1},\\
        0 & \text{otherwise.}
    \end{cases}
\end{split}
\end{align}
Here $\mathcal{Q}$ is the set of all co-visible points, including the point $\boldsymbol p$ itself, $d_j$ is the inverse depth of a point $\boldsymbol q_j$ and $L_j$ is the 2D semantic map corresponding to keyframe $j$.
$\text{dist}_{\min}$ defines a minimum object distance.
By utilizing the inverse depth $d_j$ as a weighting factor for the corresponding vote, we consider the fact that predictions for points far from the camera are generally less accurate than close points.
Meanwhile, image regions for very close objects commonly suffer from significant motion blur and strong perspective distortion, which again results in inaccurate predictions.
This is accounted for by setting a minimum distance threshold $\text{dist}_{\min}$ in the voting process.

Based on the set $V$, the point $\boldsymbol p$ gets the label with the highest weight assigned:
\begin{align}
    L_i(\boldsymbol p) \leftarrow \arg \max \left( V \right)
\end{align}

In this way, point labels are obtained in consideration of the temporal semantic segmentation information.
As the \ac{vo} front-end provides only a sparse set of points, the temporally consistent labels are also only provided for this set.

\subsection{Semantic 3D Point Cloud Generation}
The final step of our pipeline is to generate the semantically labeled 3D point cloud.
Based on the modules presented in Section~\ref{sec:semantic_segmentation} and  Section~\ref{sec:temporal_filtering}, we are able to assign semantic labels to points generated by the \ac{vo} (Sec.~\ref{sec:3d_mapping}).
Finally, these labeled 2D points are projected to 3D and transformed to a globally consistent 3D frame forming a sparse 3D semantic map as the example shown in Figure~\ref{fig:semantic_map_sample}. 

%% file: sections/experiments.tex
\section{Experiments}\label{sec:experiments}

\begin{table*}[t!]
    \centering
    \caption{Evaluation results in per-class \acs{iou} and \acs{miou} metrics of our approach on KITTI-360 dataset.}
    \vspace{1em}
    \label{tab:miou_results}
    \resizebox{\textwidth}{!}{%
    \begin{tabular}{c|c @{\hspace{1\tabcolsep}} c@{\hspace{1\tabcolsep}}c@{\hspace{1\tabcolsep}}c@{\hspace{1\tabcolsep}}c@{\hspace{1\tabcolsep}}c@{\hspace{1\tabcolsep}}c@{\hspace{1\tabcolsep}}c@{\hspace{1\tabcolsep}}c@{\hspace{1\tabcolsep}}c@{\hspace{1\tabcolsep}}c@{\hspace{1\tabcolsep}}c@{\hspace{1\tabcolsep}}c@{\hspace{1\tabcolsep}}c@{\hspace{1\tabcolsep}}c@{\hspace{1\tabcolsep}}c@{\hspace{1\tabcolsep}}c@{\hspace{1\tabcolsep}}c|c}
    \rot{0}{class} & \rot{0}{road} & \rot{0}{sidewalk} & \rot{0}{building} & \rot{0}{wall} & \rot{0}{fence} & \rot{0}{pole} & \rot{0}{traffic light} & \rot{0}{traffic sign} & \rot{0}{vegetation} & \rot{0}{terrain} & \rot{0}{car} & \rot{0}{truck} & \rot{0}{bus} & \rot{0}{train} & \rot{0}{motorcycle} & \rot{0}{bicycle} & \rot{0}{person} & \rot{0}{rider} & \rot{0}{mIoU} \\ \hline
    Baseline         & 94.4 & 79.4 & 86.6 & 54.6 & 34.4 & 39.2 & 6.6  & 45.7 & 90.1 & 60.8 & 83.4 & 71.7 & 89.1 & 91.8 & 46.3 & 19.3 & 36.2 & 4.0 & 58.5 \\
    \acs{tcl} Mono   & 95.3 & 83.6 & 87.7 & \textbf{63.3} & 42.4 & 47.5 & 9.9  & 51.2 & 91.3 & 64.9 & 83.7 & 77.6 & 92.4 & 92.3 & 55.5 & 37.9 & 64.3 & \textbf{5.3} & 63.6  \\
    \acs{tcl} Stereo & \textbf{95.4} & \textbf{84.1} & \textbf{90.9} & \textbf{63.3} & \textbf{42.7} & \textbf{51.0} & \textbf{10.1} & \textbf{52.0} & \textbf{91.8} & \textbf{65.3} & \textbf{86.1} & \textbf{79.8} & \textbf{94.3} & \textbf{95.8} & \textbf{57.4} & \textbf{38.2} & \textbf{68.2} & \textbf{5.3} & \textbf{65.1} \\ \hline
    \end{tabular}%
    }
\end{table*}

\begin{table*}[t!]
    \centering
    \caption{Evaluation results in class-wise \acs{ac} and \acs{oa} metrics of our approach on KITTI-360 dataset.}
    \vspace{1em}
    \label{tab:acc_results}
    \resizebox{\textwidth}{!}{%
    \begin{tabular}{c|c@{\hspace{1\tabcolsep}}c@{\hspace{1\tabcolsep}}c@{\hspace{1\tabcolsep}}c@{\hspace{1\tabcolsep}}c@{\hspace{1\tabcolsep}}c@{\hspace{1\tabcolsep}}c@{\hspace{1\tabcolsep}}c@{\hspace{1\tabcolsep}}c@{\hspace{1\tabcolsep}}c@{\hspace{1\tabcolsep}}c@{\hspace{1\tabcolsep}}c@{\hspace{1\tabcolsep}}c@{\hspace{1\tabcolsep}}c@{\hspace{1\tabcolsep}}c@{\hspace{1\tabcolsep}}c@{\hspace{1\tabcolsep}}c@{\hspace{1\tabcolsep}}c|c}
    \rot{0}{class} & \rot{0}{road} & \rot{0}{sidewalk} & \rot{0}{building} & \rot{0}{wall} & \rot{0}{fence} & \rot{0}{pole} & \rot{0}{traffic light} & \rot{0}{traffic sign} & \rot{0}{vegetation} & \rot{0}{terrain} & \rot{0}{car} & \rot{0}{truck} & \rot{0}{bus} & \rot{0}{train} & \rot{0}{motorcycle} & \rot{0}{bicycle} & \rot{0}{person} & \rot{0}{rider} & \rot{0}{OA} \\ \hline
    baseline         & 96.4 & 88.9 & 94.4 & 67.7 & 56.1 & 82.2 & 81.3 & 80.3 & 94.2 & 83.6 & 96.6 & 87.0 & 95.4 & 97.3 & 52.8 & 94.5 & 90.9 & \textbf{100.0} & 91.4 \\
    \acs{tcl} Mono   & 97.3 & 91.2 & 94.7 & 73.0 & 58.1 & 83.8 & 87.3 & 83.2 & 95.5 & \textbf{85.8} & 97.9 & 93.3 & 96.3 & 97.3 & 62.4 & 95.8 & 94.0 & \textbf{100.0} & 94.1 \\
    \acs{tcl} Stereo & \textbf{97.5} & \textbf{91.9} & \textbf{95.1} & \textbf{73.2} & \textbf{59.9} & \textbf{84.5} & \textbf{87.6} & \textbf{83.4} & \textbf{95.6} & \textbf{85.8} & \textbf{98.2} & \textbf{93.4} & \textbf{98.5} & \textbf{98.9} & \textbf{64.8} & \textbf{97.0} & \textbf{94.6} & \textbf{100.0} & \textbf{95.2} \\ \hline
    \end{tabular}%
    }
\end{table*}

In the following, we evaluate our pipeline on the KITTI-360 dataset~\cite{xie2016kitti360}.
We propose the generation of improved ground truth semantic labels for the generated 3D points by fusing labels of 2D images and 3D \ac{lidar} points from the KITTI-360 dataset. 
We show both quantitative and qualitative results and demonstrate the performance improvement with the temporally consistent labeling.
In addition to the systematic evaluations on the KITTI-360 dataset, Figure~\ref{fig:berlin_map} shows the capability of our pipeline to generate a city-scale semantic map from data collected by a fleet of vehicles.

\subsection{Dataset}

KITTI-360~\cite{xie2016kitti360} is a large-scale dataset that covers \SI{74}{km} suburban traffic scenes with over 400k images and over 100k \ac{lidar} scans. It provides rich annotations, including manually labeled 3D \ac{lidar} point clouds and front-view 2D semantic labels transferred from the 3D annotations.

As described in Section \ref{sec:3d_mapping}, our 3D mapping core generates sparse point clouds only from images \cite{wang2017stereoDSO}.
The 3D points generated by our pipeline differ from the \ac{lidar} points provided in the KITTI-360 dataset.
Hence, it is necessary to associate our generated points with the ground truth provided by the KITTI-360 dataset.
Then we can properly evaluate the performance of our semantic mapping approach.

\begin{figure}[t]
    \centering
    \includegraphics[width=0.48\textwidth]{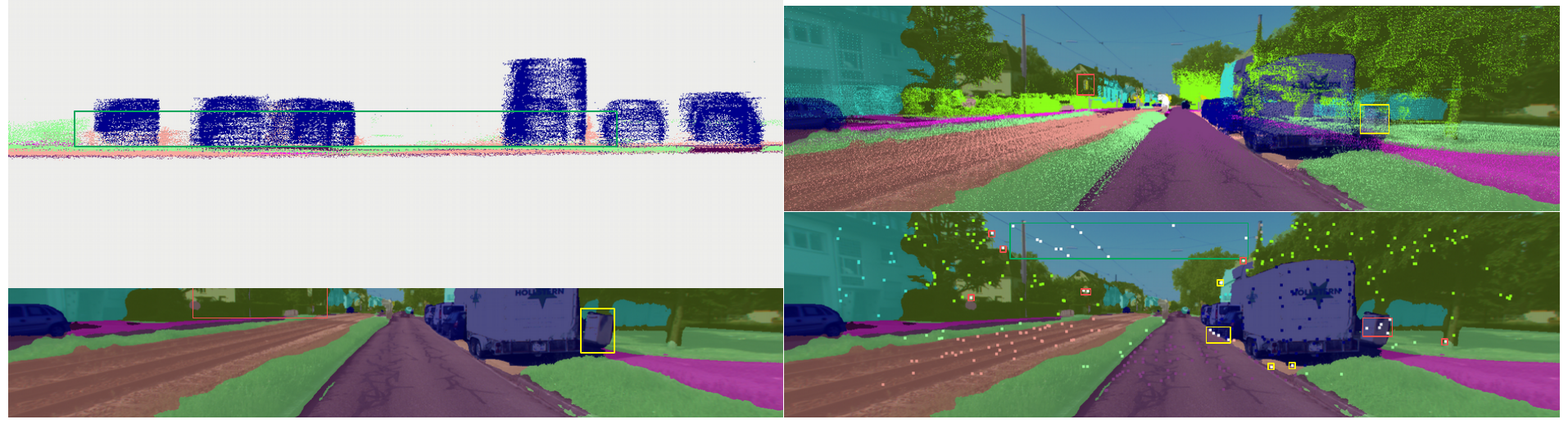}
    \vspace{-1.em}
    \caption{\textbf{Label quality} of 3D and 2D semantic ground truth of the KITTI-360 dataset \cite{xie2016kitti360} and the semantic labels we associate for our point clouds. \textbf{\textit{Top Left:}} In the \textit{green} box, the points around the cars \textit{in blue} are mislabeled as parking class \textit{in pink} in KITTI-360 3D semantic ground truth. \textbf{\textit{Bottom Left:}} In the \textit{red} box, the building and traffic sign are mislabeled as vegetation and in the \textit{yellow} box, a mail box is mislabeled as a car in KITTI-360 2D semantic ground truth. \textbf{\textit{Top Right:}} The 3D semantic labels given by KITTI-360 projected into one keyframe where the traffic light and the mail box are not mislabeled. \textbf{\textit{Bottom Right:}} The ground truth labels we associated for our point clouds based on the given KITTI-360 2D and 3D labels. The filtered points in \textit{white}: the mislabeled points in \textit{red} boxes, the points on the boundaries in \textit{yellow} boxes, and the points in the sky in the \textit{green} box.}
    \label{fig:label_quality}
\end{figure}

Our pipeline generates point clouds solely from image data without integrating global measurements like \ac{gnss}.
Therefore, we cannot guarantee that our 3D model can be globally aligned with the provided \ac{lidar} points up to centimeter accuracy, even though the \ac{vo} front-end provides locally highly accurate camera poses and 3D reconstructions.
Besides, the distribution of the points reconstructed by the \ac{vo} is generally non-isotropic and highly dependent on the distance of an object point to the camera.
Therefore, instead of performing global point cloud alignment, we use the provided ground truth poses to perform a keyframe-wise alignment between the camera-based reconstruction and the \ac{lidar} ground truth.
For all points hosted in a keyframe, we perform point matching in 2D by projecting the labeled \ac{lidar} points into the keyframe image and assigning the labels of \ac{lidar} points to the matched points. We utilize \ac{lidar} points up to 100 meters away from the camera, which makes the projected points dense enough in 2D to have all points match a label. The top right plot in Figure~\ref{fig:label_quality} shows the density of the projected 3D points.
Additionally, depth information is utilized to reject false assigned point pairs.

In KITTI-360 dataset, the labels of the 3D points are generated by placing primitive cuboids and ellipsoids around objects and assigning a class label to the enclosed points. Besides, it focuses on static scene objects.
As a result, not all points are accurately labeled in 3D. The given 2D semantic ground truth is derived from the 3D labels, making the 2D ground truth not absolutely accurate as well. The left column of Figure~\ref{fig:label_quality} shows two examples that have mislabeled objects.
Therefore, we fuse both the 2D image labels and the 3D \ac{lidar} labels provided by the dataset to obtain reliable ground truth.
Only if the labels of both 2D and projected 3D points are consistent, they are considered as ground truth for evaluation, as shown in the bottom right plot in Figure~\ref{fig:label_quality}.

\subsection{Evaluation}

\subsubsection{Settings}
We evaluate our approach on all 9 released sequences of the KITTI-360 dataset. The evaluation is done on all classes as defined except the void-type classes, e.g. \textit{unlabeled} and \textit{ego vehicle}, etc., and the \textit{sky} class as our map does not contain points representing the sky. The performance is evaluated for the baseline model presented in Section~\ref{sec:semantic_segmentation}~\cite{zhu2019improving} and the \ac{tcl} module based on the pre-trained network weights\footnote[1]{\href{https://github.com/NVIDIA/semantic-segmentation/tree/sdcnet}{https://github.com/NVIDIA/semantic-segmentation/tree/sdcnet}}.
For the \ac{tcl} module we provide results using semantic labels predicted only for the left camera images (Mono) as well as using semantic labels for both left and right images (Stereo).

\subsubsection{Measurement Metric}
To evaluate the performance of semantic label prediction, we follow the standard to apply per-class \acf{iou} \cite{everingham2015} and the \acf{miou} metrics over all sequences. 
\begin{align}
\begin{split}
    IoU_c = \frac{TP_c}{TP_c + FP_c + FN_c}
\end{split}
\begin{split}
    mIoU = \frac{1}{C}\sum_{c=1}^C{IoU_c}
\end{split}
\end{align}
where $TP_c$, $FP_c$, and $FN_c$ indicate the number of true positive, false positive, and false negative estimations for each class $c$ in all sequences respectively.
$C$ is the number of evaluation classes.

As the number of points of each class varies prominently, we also measure the class-wise prediction \acf{ac} and the \acf{oa} over all sequences:
\begin{align}
\begin{split}
    mAcc = \frac{TP_c}{N_c}
\end{split}
\begin{split}
    OA = \frac{\sum_{c=1}^C TP_c} {\sum_{c=1}^C N_c}
\end{split}
\end{align}
where $N_c$ is the number of points predicted as class $c$.

\begin{figure}[t!]
    \centering
    \includegraphics[width=0.98\linewidth]{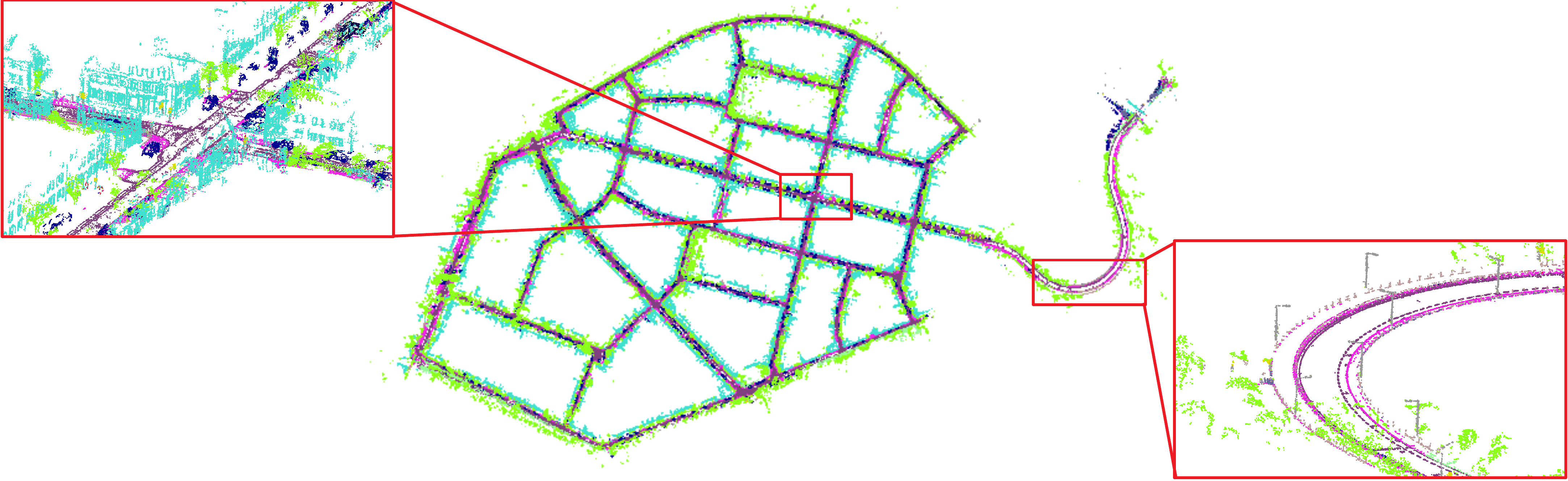}
    \caption{\textbf{Large-scale semantic map} generated by the proposed pipeline. The figure shows the entire globally consistent reconstruction of Sequence~9 from the KITTI-360 dataset~\cite{xie2016kitti360} and zoomed-in sections show the fine-grained reconstruction details.}
    \label{fig:large_scale_map}
\end{figure}

\subsubsection{Results}

\begin{figure*}[t!]
    \centering
    \vspace{0.2em}
    \includegraphics[width=0.99\textwidth]{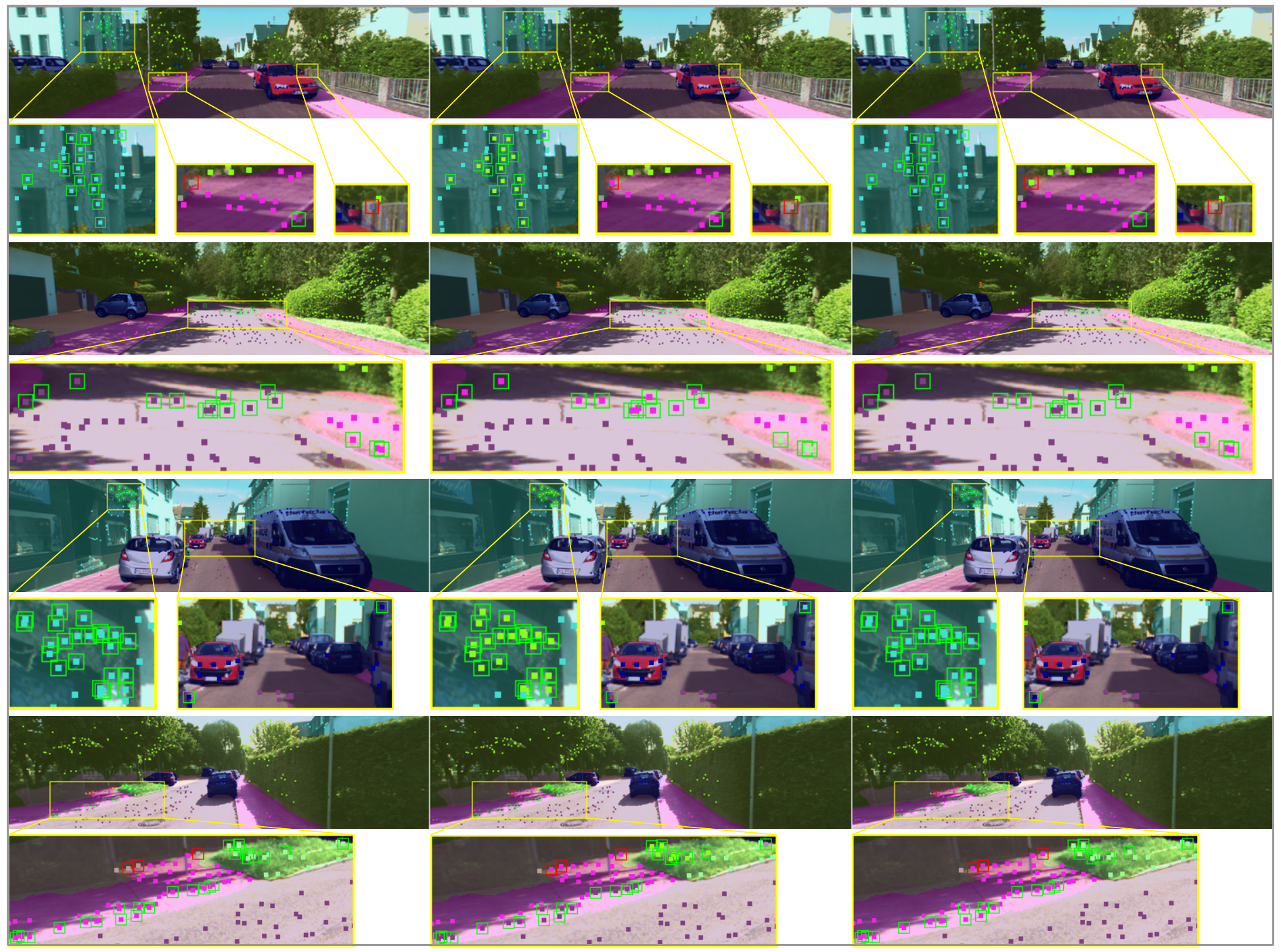}
    \caption{\textbf{Semantic point labels generated by our approach.} The sparse points hosted in a keyframe are overlaid with the 2D ground truth labels and the image from KITTI-360. This is only a subset of all points in the optimization window.
    Two rows represent one example, where the first and second row shows the prediction results in full images and the zoom-in windows, respectively.
    Columns show \textbf{\textit{(left)}} sparse ground truth labels, \textbf{\textit{(center)}} labels predicted by the baseline model, \textbf{\textit{(right)}} labels estimated by \acs{tcl}, respectively.
    Points which are wrongly predicted by the baseline but corrected by the \acs{tcl} are highlighted with \textit{green} boxes.
    Points not corrected by the \acs{tcl} are highlighted with \textit{red} boxes.}
    \label{fig:pred_results}
\end{figure*}

Table~\ref{tab:miou_results} shows the evaluation results in per-class \ac{iou} and \ac{miou}.
Table~\ref{tab:acc_results} presents them in class-wise \ac{ac} and \ac{oa}. 
These class-wise measurements indicate that our approach performs well on primary static classes, including road, sidewalk, building and vegetation classes, and also on some dynamic classes, car, truck, and bus because most objects of these classes are not moving in these sequences. When comparing Table~\ref{tab:miou_results} and Table~\ref{tab:acc_results}, we can conclude that our approach performs better in terms of precision than recall, especially on classes representing small objects, e.g. traffic sign, traffic light, bicycle, person. These small objects have more complex and elongated boundaries, and the points near boundaries are more challenging to be perfectly classified. Besides, there are only a few points corresponding to the small objects, and thus a wrong prediction can reduce the \ac{iou} and \ac{ac} by a large margin. Thus, the large classes has better results than the small classes.

The proposed \acf{tcl} fixes inconsistent predictions which mainly occur on object boundaries across consecutive frames.
Therefore, \ac{tcl} outperforms the baseline model, especially on small objects, e.g. pole, motorcycle, bicycle, and person.
Furthermore, the stereo version clearly outperforms the mono version.

Qualitative results of semantic point cloud labels generated by our approach are shown in Figure~\ref{fig:pred_results}. It illustrates how \ac{tcl} can improve the label estimation accuracy.
Additionally, Figure~\ref{fig:large_scale_map} shows an entire large-scale map generated by our approach from the sequence 9 in KITTI-360 dataset.

%% file: sections/conclusion.tex
\section{Conclusion}\label{sec:conclusion}

We presented a large-scale semantic mapping pipeline, which combines a state-of-the-art direct sparse \ac{vo} front-end and a back-end with global optimization and image-based semantic segmentation.
We demonstrated that predictions of the semantic segmentation network could be improved by incorporating temporal correspondences established by \ac{vo}.
Furthermore, we showed that our pipeline is capable of generating city-scale semantic maps covering thousands of kilometers of the road by using a fleet of vehicles.
Such large-scale semantic maps can serve as intermediate results towards fully vectorized HD maps.
Besides, we see it as a next step to build semantic 3D volumetric maps at scale with the combination of state-of-the-art dense reconstruction approaches like~\cite{wimbauer2020monorec}.

%% file: sections/acknowledgement.tex
\section*{Acknowledgement}

We thank the entire Artisense team for their support in setting up the pipeline and making this project happen. 